\documentclass{article} 

\usepackage[preprint]{neurips_2024}

\usepackage{hyperref}
\usepackage{url}
\usepackage[utf8]{inputenc} 
\usepackage[T1]{fontenc}    
\usepackage{hyperref}       
\usepackage{booktabs}       
\usepackage{amsfonts}       
\usepackage{nicefrac}       
\usepackage{microtype}      
\usepackage{xcolor}         
\usepackage{multirow}
\usepackage{graphicx}
\usepackage{lscape}
\usepackage{booktabs}
\usepackage{wrapfig,lipsum,booktabs}
\usepackage{makecell}
\usepackage{amsmath}
\usepackage{graphicx}
\usepackage{subcaption}
\usepackage{enumitem}

\usepackage{algorithm}
\usepackage{csquotes}
\usepackage{algorithmicx}
\usepackage{algpseudocode}
\usepackage{bm}
\usepackage{amsmath}  
\usepackage[normalem]{ulem}
\allowdisplaybreaks[4]
\newcommand{\method}{ProtoMol}

\title{ProtoMol: Enhancing Molecular Property Prediction via Prototype-Guided Multimodal Learning}

\author{%
  Yingxu Wang\thanks{Equal Contributions.}\\
  MBZUAI\\
  \texttt{yingxv.wang@gmail.com} \\
  \And
  Kunyu Zhang\footnotemark[1]\\
  University of Zhengzhou\\
  \texttt{kunyu.zky@gmail.com}
  \And
  Jiaxin Huang \\
  MBZUAI\\
  \texttt{jiaxin.huang@mbzuai.ac.ae} 
  \And
  Nan Yin \\
  HKUST\\
  \texttt{yinnan8911@gmail.com} 
  \And
  Siwei Liu\thanks{Corresponding author.} \\
  University of Aberdeen\\
  \texttt{siwei.liu@abdn.ac.uk} 
  \And
  Eran Segal\footnotemark[2] \\
  MBZUAI, Weizmann Institute of Science\\
  \texttt{eran.segal@weizmann.ac.il}
}

\begin{document}

\maketitle
\begin{abstract}

Multimodal molecular representation learning, which jointly models molecular graphs and their textual descriptions, enhances predictive accuracy and interpretability by enabling more robust and reliable predictions of drug toxicity, bioactivity, and physicochemical properties through the integration of structural and semantic information. However, existing multimodal methods suffer from two key limitations: (1) they typically perform cross-modal interaction only at the final encoder layer, thus overlooking hierarchical semantic dependencies; (2) they lack a unified prototype space for robust alignment between modalities. To address these limitations, we propose \method{}, a prototype-guided multimodal framework that enables fine-grained integration and consistent semantic alignment between molecular graphs and textual descriptions. \method{} incorporates dual-branch hierarchical encoders, utilizing Graph Neural Networks to process structured molecular graphs and Transformers to encode unstructured texts, resulting in comprehensive layer-wise representations. Then, \method{} introduces a layer-wise bidirectional cross-modal attention mechanism that progressively aligns semantic features across layers. Furthermore, a shared prototype space with learnable, class-specific anchors is constructed to guide both modalities toward coherent and discriminative representations. Extensive experiments on multiple benchmark datasets demonstrate that \method{} consistently outperforms state-of-the-art baselines across a variety of molecular property prediction tasks.
\end{abstract}

\section{Introduction}

Accurate molecular representation is fundamental to a broad spectrum of applications in bioinformatics, such as drug discovery, toxicity assessment, and functional genomics~\cite{ mcgibbon2024intuition, lipinski1997experimental}. In recent years, multimodal learning has become an effective approach for integrating diverse molecular information, especially for predicting biologically relevant properties of small molecules and biomacromolecules~\cite{wang2025bridging, radford2021learning}. In typical bioinformatics pipelines, molecular data are represented in two complementary forms: structured graphs~\cite{wang2024chain, duvenaud2015convolutional}, which depict atomic connectivity and chemical bonding, and descriptive textual annotations~\cite{smiles, flores2024systematic,wang2025dynamically}, such as SMILES strings and expert summaries, that capture chemical, functional, or pharmacological characteristics. Graph-based models provide a faithful depiction of molecular structure, effectively capturing spatial interactions and stereochemical configurations essential for understanding biological function~\cite{schutt2017schnet, zhou2023uni, zheng2024predicting}. In contrast, textual descriptors provide a compact, information-rich sequence that can highlight key functional groups, bioactive motifs, and provide higher-level semantic abstraction within biological or pharmacological contexts~\cite{gomez2018automatic, chithrananda2020chemberta, sadeghi2024can}. By combining these modalities, it becomes possible to develop more comprehensive and interpretable predictive models, thereby advancing molecular property prediction and supporting hypothesis-driven biomedical research~\cite{liu2023multi, rong2020self,shao2025enhanced}.

Early approaches in molecular representation learning primarily relied on textual descriptors to capture chemical information. Among these, the Simplified Molecular Input Line Entry System (SMILES)~\cite{smiles} emerged as a widely adopted format for encoding molecular structures as linear sequences of characters. This sequential representation enables the use of established sequence modeling architectures, including recurrent neural networks (RNNs)~\cite{elman1990finding}, convolutional neural networks (CNNs)~\cite{lecun2002gradient}, and Transformers~\cite{vaswani2017attention, yoshikai2024difficulty, guo2023can,wang2023cl4ctr}, which leverage the syntactic regularity and semantic compactness of SMILES to extract features such as atom types, bond orders, and functional groups~\cite{ucak2023improving, tang2025molfcl, rogers2010extended}. While SMILES provides a concise and expressive encoding, its inherent linear structure limits the preservation of molecular topology, making it difficult to accurately represent complex structural features, including stereochemistry, ring closures, and branching patterns that are essential for understanding biological function and pharmacological properties~\cite{ganeeva2024chemical, ucak2023improving, krenn2020self}. To overcome this limitation, graph-based models have been developed, where molecules are represented as graphs with atoms as nodes and chemical bonds as edges~\cite{zhao2024molecular, boulougouri2024molecular, wang2025nested}. Graph Neural Networks (GNNs)~\cite{zhou2023uni, wang2024chain, ju2024survey,yin2023coco} enable the propagation of information through message-passing mechanisms, allowing for the aggregation of both local atomic environments and long-range structural dependencies~\cite{zhou2023uni, zheng2024predicting,wang2024dusego}. This approach preserves spatial connectivity and topological relationships, providing a more biologically relevant representation compared to sequence-based methods~\cite{schutt2018schnet, zheng2024predicting}. Recent research has increasingly focused on integrating textual and graph-based information to achieve more comprehensive and informative molecular representations~\cite{liu2023multi, moayedpour2024representations}. These multimodal approaches align the structural fidelity of molecular graphs with the semantic richness of textual descriptions through cross-modal interactions, joint embedding spaces, or aligned training objectives, thereby enabling the capture of a broader spectrum of molecular characteristics~\cite{liu2023multi, wang2023molecular}. To further enhance semantic consistency and model interpretability, prototype-based learning strategies have been introduced, where learnable, class-specific prototypes serve as semantic anchors to align heterogeneous representations from graphs and text~\cite{he2024prototype, zhang2024molfescue, snell2017prototypical}. Prototype-guided frameworks have demonstrated significant potential for improving the discriminative power and generalizability of molecular embeddings, especially in tasks such as few-shot property prediction, identification of rare or complex substructures, and discovery of chemically or biologically relevant functional groups~\cite{hou2024attribute, seo2023interpretable, chen2019looks}. As a result, these advances in multimodal and prototype-guided learning are driving progress in computational drug discovery, functional genomics, and other key areas of biomedical research.


However, current multimodal molecular representation learning methods still face two fundamental limitations that restrict their applicability in real-world bioinformatics and biomedical research. (1) Most existing approaches perform cross-modal interaction only at the final layer of graph and textual encoders, overlooking the hierarchical alignment across intermediate semantic levels~\cite{han2025integrated, lu2019vilbert}. Molecular graphs naturally encode atomic connectivity and spatial topology, while textual descriptions provide functional and pharmacological annotations in sequential form~\cite{zhang2024mvmrl, morgan1965generation}. These modalities not only differ in representational structure and semantic density, but also encapsulate distinct aspects of biological relevance. Restricting their interaction to a single, final layer of encoders limits the model’s ability to capture fine-grained, context-dependent correspondences that are crucial for tasks such as bioactivity prediction~\cite{rollins2024molprop}.  Without intermediate fusion, biologically relationships between modalities remain uncaptured, ultimately diminishing the interpretability and predictive power of the model in real-world biomedical scenarios. (2) Although prototype learning has been incorporated into multimodal molecular representation learning, most existing approaches still construct prototypes independently for each modality~\cite{fan2023pmr, vinyals2016matching}. Rather than establishing a unified prototype space that integrates both structural and semantic information following cross-modal fusion, these methods rely on separate, modality-specific prototype spaces. This decoupled design prevents the formation of consistent and biologically meaningful semantic anchors across modalities, which often results in distributional shifts and semantic misalignment even for the same molecular entity~\cite{wang2023connecting, frome2013devise}. As a result, such models may struggle to reliably identify or annotate functionally important substructures, rare chemotypes, or disease-associated motifs~\cite{amara2023explaining, bemis1996properties}. These elements are crucial for a variety of bioinformatics applications, including drug discovery and biomarker identification. Without a unified semantic space to reconcile complementary molecular information, the resulting representations tend to be fragmented and poorly generalizable. This ultimately limits the effectiveness of these models in supporting translational bioinformatics and precision medicine.

To address these limitations, we propose \method{}, a prototype-guided multimodal framework specifically designed to enable fine-grained integration and unified semantic alignment between molecular graphs and their associated textual descriptions. Our framework is built on a dual-branch hierarchical encoder architecture, with a multi-layer GNNs capturing the structural and topological characteristics of molecules, and a Transformer-based encoder processing rich functional and pharmacological annotations from molecular texts. The resulting layer-wise semantic representations effectively reflect both the chemical connectivity and the biological context of each molecule. To facilitate cross-modal understanding, we introduce a layer-wise bidirectional cross-modal attention mechanism, which progressively aligns features from both modalities at multiple semantic levels. Moreover, we establish a unified semantic prototype space composed of learnable, class-specific, and modality-invariant prototypes, serving as consistent semantic anchors across modalities. After cross-modal fusion, both graph and text representations are projected into this shared prototype space, where their alignment is supervised using a dual-objective training approach. This approach combines a prototype alignment loss, which enforces consistency between modalities based on Kullback-Leibler divergence~\cite{zhuang2025math, zhang2024dual, li2020prototypical}, with a prototype contrastive loss that enhances intra-class compactness and inter-class separability. To verify the effectiveness of \method{}, we conduct extensive experiments on a range of molecular property prediction benchmarks. The results show that \method{} consistently outperforms current state-of-the-art approaches in most cases.

Our main contributions are as follows:
 \begin{itemize}
 \item We propose \method{}, a prototype-guided multimodal framework for multi-modal molecular representation learning that jointly captures both structural and semantic information from molecular graphs and their corresponding textual descriptions, leading to improved predictive performance and enhanced interpretability.
 \item \method{} integrates dual-branch hierarchical encoders with a novel layer-wise bidirectional cross-modal attention mechanism and a unified semantic prototype space, enabling fine-grained semantic alignment and consistent, modality-invariant representation learning through prototype-level supervision.
 \item We demonstrate the practical effectiveness of \method{} through extensive experiments on diverse molecular property prediction benchmarks, showing that it consistently outperforms state-of-the-art baselines in most cases. 
 \end{itemize}

\section{Methodology}

\subsection{Preliminary}
Given a molecular graph $G = (V, E, \mathbf{X})$, where $V$ denotes the set of atoms, $E$ represents the set of chemical bonds, and $\mathbf{X}$ is the atom feature matrix, together with a textual description $\mathbf{t}$ derived from the molecule’s SMILES representation, we address the task of molecular property prediction. This task includes both classification and regression settings. For classification, the objective is to predict binary or multi-class molecular attributes, such as the toxicity or biological activity of molecules. For regression, the goal is to estimate continuous-valued properties, including the solubility and lipophilicity of molecules. The proposed \method{} method processes each molecular instance as a multimodal input pair $(G, \mathbf{t})$ and outputs a prediction $\hat{y} \in \mathcal{Y}$, where $\mathcal{Y}$ corresponds to either a discrete set of class labels or a continuous property value.

\begin{figure*}[t]
    \centering
    \includegraphics[width=1.0\linewidth]{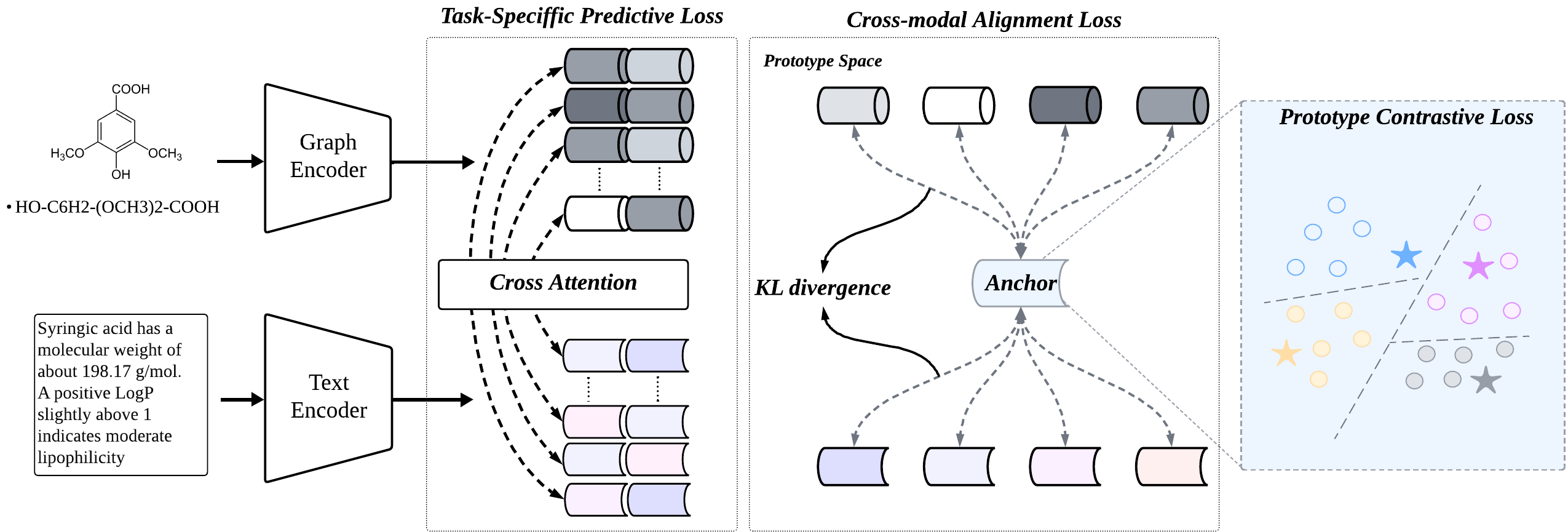}
    \caption{
    Overall architecture of the proposed \method{}. The workflow begins by independently encoding molecular graphs and their corresponding textual descriptions using hierarchical GNN and Transformer-based encoders, respectively. At each representation layer, bidirectional cross-modal attention modules facilitate information exchange between the graph and text modalities, progressively enriching the learned embeddings. These hierarchical representations are then projected into a unified semantic prototype space, which acts as a shared anchor point for both modalities. At multiple levels, prototype-based contrastive learning and alignment objectives are applied to refine the representations, enhancing intra-class compactness and inter-class separability across modalities. }
    \label{fig:overview}
    \vspace{-0.3cm}
\end{figure*}

\subsection{Overview}

In this section, we present \method{}, a prototype-guided multimodal framework designed to integrate molecular graphs and their corresponding textual descriptions for accurate molecular property prediction, as shown in Figure \ref{fig:overview}. The overall architecture comprises three key components: (1) a dual-branch encoder, which captures hierarchical representations from both graph and text modalities using a multi-layer graph neural network and a Transformer, respectively; (2) a structured cross-modal interaction module, which employs layer-wise bidirectional attention to model semantic dependencies between modalities at multiple hierarchical levels; and (3) a unified semantic prototype space that acts as modality-invariant anchors to facilitate robust cross-modal alignment. By jointly modeling hierarchical interactions and prototype-level alignment, \method{} enables fine-grained semantic fusion and promotes the learning of consistent and discriminative representations across modalities.

\subsection{Dual-Branch Hierarchical Representation Learning}

To effectively capture both structural and semantic information from molecular graphs and their associated textual descriptions, the proposed \method{} framework employs a dual-branch encoder that independently processes each modality within a hierarchical architecture.

\subsubsection{Graph Branch for Structural Representation Learning.} Given a molecular graph $G$, we employ a multi-layer Graph Neural Networks (GNNs) to encode both local structural features and global topological dependencies. Let $\mathbf{h}_v^{(l)} \in \mathbb{R}^{d_g}$ denote the representation of atom $v \in V$ at the $l$-th GNN layer, $d_g$ denotes the dimensionality of the node representation. The initial representation of atom $v$ is obtained via a learnable multi-layer perceptron (MLP):
\begin{equation}
\mathbf{h}_v^{(0)} = \text{MLP}(\mathbf{x}_v),
\end{equation}
where $\mathbf{x}_v \in \mathbb{R}^{d}$ denotes the raw input feature of atom $v$, $d$ is the dimensionality of raw input feature. The GNN then iteratively updates each atom representation through a message-passing mechanism aggregating information from its local neighborhood~\cite{xu2018powerful,wang2025nested,wang2024degree}. At the $l$-th layer, the representation $\mathbf{h}_v^{(l)}$ of atom $v$ is computed as:
\begin{align}
\mathbf{h}_{v}^{(l)} &= \mathcal{C}^{(l-1)} \left( \mathbf{h}_{v}^{(l-1)},\ 
\mathcal{A}^{(l-1)}\left( \left\{ \mathbf{h}_{u}^{(l-1)} \mid u \in \mathcal{N}(v) \right\} \right) \right), \nonumber \\
&\qquad \qquad \qquad l = 1, \dots, L,
\end{align}
where $\mathcal{A}^{(l-1)}(\cdot)$ is the aggregation function (e.g., summation, mean, or attention-based pooling) over the neighbors $\mathcal{N}(v)$ of atom $v$, and $\mathcal{C}^{(l-1)}(\cdot)$ is the combination function that integrates the aggregated neighborhood information with the current atom representation~\cite{wang2024sgac,wang2024dusego}.

To capture hierarchical structural information from molecular graphs, we apply a readout operation at each layer of the graph neural network to obtain a series of graph-level representations~\cite{xu2018powerful,chen2023heterogeneous,yin2025dream}. Specifically, at the $l$-th layer, the graph-level representation is defined as
\begin{equation}
\mathbf{z}_g^{(l)} = \text{READOUT}\left( \left\{ \mathbf{h}_v^{(l)} \mid v \in V \right\} \right),
\end{equation}
where $\text{READOUT}(\cdot)$ denotes a permutation-invariant function such as summation, mean, or max pooling~\cite{xu2018powerful,yin2024dream,yao2023improving}. This results in a set of hierarchical graph-level representations:
\begin{equation*}
\mathbf{Z}_g = \left \{ \mathbf{z}_g^{(1)}, \dots, \mathbf{z}_g^{(L)} \right \}, \quad \mathbf{z}_g^{(l)} \in \mathbb{R}^{d_g}.
\end{equation*}

\subsubsection{Text Branch for Semantic Representation Learning.}
Given a textual description $\mathbf{t} = [t_1, t_2, \dots, t_n]$ consisting of $n$ tokens corresponding to a molecular graph $G$ (for example, \enquote{a benzene ring connected to a hydroxyl group and a carboxylic acid}), we leverage a pretrained large language model, such as Qwen~\cite{yang2025qwen3}, to generate context-aware semantic embeddings for the textual input. Pretrained language models are capable of capturing nuanced semantic relationships, contextual dependencies, and domain-specific chemical terminology by drawing upon knowledge acquired from large-scale text corpora~\cite{grattafiori2024llama, yang2025qwen3}. Specifically, each token $t_i$ is mapped to a dense, context-dependent embedding vector $\mathbf{e}_i \in \mathbb{R}^{d_t}$ as follows:
\begin{equation}
\mathbf{e}_i = f_{\text{text}}(t_i), \quad i = 1, \dots, n,
\end{equation}
where $d_t$ denotes the dimensionality of the token embedding and $f_{\text{text}}(\cdot)$ represents the pretrained language model. To obtain an initial global representation that summarizes the overall semantic content of the molecular description, we apply mean pooling across all token embeddings:
\begin{equation}
\mathbf{z}_t^{(0)} = \frac{1}{n} \sum_{i=1}^{n} \mathbf{e}_i,
\end{equation}
where $\mathbf{z}_t^{(0)} \in \mathbb{R}^{d_t}$ denotes the aggregated, context-aware representation of the textual input, effectively capturing both local token-level semantics and the broader molecular context.

To capture hierarchical semantic information, we further refine the initial text representation $\mathbf{z}_t^{(0)}$ using a Transformer composed of $L$ layers. At each layer, the representation is updated according to
\begin{equation}
\mathbf{z}_t^{(l)} = \mathrm{Transformer}^{(l)}(\mathbf{z}_t^{(l-1)}), \quad l = 1, \dots, L.
\end{equation}
This process yields a set of hierarchical text-level representations:
\begin{equation*}
\mathbf{Z}_t = \left\{ \mathbf{z}_t^{(1)}, \dots, \mathbf{z}_t^{(L)} \right\}, \quad \mathbf{z}_t^{(l)} \in \mathbb{R}^{d_t}.
\end{equation*}

\subsection{Layer-wise Bidirectional Attention for Cross-Modal Interaction}

{To capture fine-grained semantic dependencies between molecular graphs and their textual descriptions, we design a hierarchical bidirectional cross-modal attention module that enables progressive structural–semantic alignment between modalities. Different from standard cross-attention mechanisms that perform a single-stage fusion at the final layer~\cite{han2025integrated,rollins2024molprop}, our design introduces layer-wise bidirectional interaction between the hierarchical representations} $\mathbf{Z}_g = \{ \mathbf{z}_g^{(1)}, \dots, \mathbf{z}_g^{(L)} \}$ and $\mathbf{Z}_t = \{ \mathbf{z}_t^{(1)}, \dots, \mathbf{z}_t^{(L)} \}$. At each layer $l$, the graph representation $\mathbf{z}_g^{(l)} \in \mathbb{R}^{d_g}$ is projected into the text representation space using a learnable transformation matrix $\mathbf{W}_{g \rightarrow t}^{(l)} \in \mathbb{R}^{d_t \times d_g}$:
\begin{equation}
{\mathbf{z}}_{g \rightarrow t}^{(l)} = \mathbf{W}_{g \rightarrow t}^{(l)} \mathbf{z}_g^{(l)}.
\end{equation}
Conversely, the text representation $\mathbf{z}_t^{(l)} \in \mathbb{R}^{d_t}$ is projected into the graph representation space via a learnable transformation matrix $\mathbf{W}_{t \rightarrow g}^{(l)} \in \mathbb{R}^{d_g \times d_t}$:
\begin{equation}
{\mathbf{z}}_{t \rightarrow g}^{(l)} = \mathbf{W}_{t \rightarrow g}^{(l)} \mathbf{z}_t^{(l)}.
\end{equation}

{We then compute the layer-wise bidirectional attention from graph to text 
by using ${\mathbf{z}}_{g \rightarrow t}^{(l)}$ as the key and value, 
and $\mathbf{z}_t^{(l)}$ as the query:
\begin{equation}
\tilde{\mathbf{z}}_t^{(l)} =
\text{CrossAttn}_{g \rightarrow t}^{(l)}
\left(
\mathbf{z}_t^{(l)},
{\mathbf{z}}_{g \rightarrow t}^{(l)},
{\mathbf{z}}_{g \rightarrow t}^{(l)}
\right),
\label{eq:g2t_crossattn_main}
\end{equation}
{where the cross-attention operation is defined as}:
\begin{equation}
\text{CrossAttn}
\left(
\mathbf{z}_t^{(l)},
{\mathbf{z}}_{g \rightarrow t}^{(l)},
{\mathbf{z}}_{g \rightarrow t}^{(l)}
\right)
=
\text{softmax}\!\left(
\frac{
\mathbf{z}_t^{(l)} 
\left({\mathbf{z}}_{g \rightarrow t}^{(l)}\right)^{\top}
}{
\sqrt{d_g}
}
\right)
{\mathbf{z}}_{g \rightarrow t}^{(l)}.
\label{eq:g2t_crossattn_def}
\end{equation}
Analogously, we compute the cross-attention from text to graph as:
\begin{equation}
\tilde{\mathbf{z}}_g^{(l)} =
\text{CrossAttn}_{t \rightarrow g}^{(l)}
\left(
\mathbf{z}_g^{(l)},
{\mathbf{z}}_{t \rightarrow g}^{(l)},
{\mathbf{z}}_{t \rightarrow g}^{(l)}
\right).
\label{eq:t2g_crossattn}
\end{equation}}

The final cross-modally enhanced representations for the graph and text modalities are obtained by applying a residual connection with the original modality-specific representations:
\begin{equation}
\mathbf{\hat{z}}_t^{(l)} = \mathbf{z}_t^{(l)} + \tilde{\mathbf{z}}_t^{(l)}, \quad \mathbf{\hat{z}}_g^{(l)} = \mathbf{z}_g^{(l)} + \tilde{\mathbf{z}}_g^{(l)}.
\end{equation}

{This layer-wise bidirectional attention mechanism facilitates progressive and structured fusion between graph and text semantics. In this process, lower layers capture structural dependency transfer, while higher layers progressively refine semantic alignment. Through this hierarchical interaction, the module effectively bridges the representational gap between structural and linguistic modalities, thereby establishing a solid foundation for unified prototype learning in the subsequent stage.}

\subsection{Modality-Invariant Prototype Space for Cross-Modal Alignment}

To bridge the semantic gap between graph and textual modalities, we construct a unified semantic prototype space that serves as a set of modality-invariant anchors for cross-modal alignment. This space consists of a collection of learnable, class-specific prototypes defined as
\begin{equation*}
\mathcal{P} = \left\{ \mathbf{p}_{c,n} \right\}_{c=1,n=1}^{C,N} \in \mathbb{R}^{C \times N \times d_p},
\end{equation*}
where each prototype $\mathbf{p}_{c,n} \in \mathbb{R}^{d_p}$ represents the $n$-th semantic anchor for class $c$ in the shared latent space. Here, $C$ denotes the number of classes in the dataset, $N$ is the number of prototypes for per class, and $d_p$ is the dimensionality of each prototype vector~\footnote{For regression tasks, we set $C = 1$ to reflect the continuous output space; for classification tasks, $C$ equals the number of discrete class labels.}.

Given the cross-modally enhanced representations $\hat{\mathbf{Z}}_g = \{ \hat{\mathbf{z}}_g^{(1)}, \dots, \hat{\mathbf{z}}_g^{(L)} \}$ and $\hat{\mathbf{Z}}_t = \{ \hat{\mathbf{z}}_t^{(1)}, \dots, \hat{\mathbf{z}}_t^{(L)} \}$, we first aggregate these representations using mean pooling:
\begin{equation}
\bar{\mathbf{z}}_g = \frac{1}{L} \sum_{l=1}^{L} \hat{\mathbf{z}}_g^{(l)}, \quad
\bar{\mathbf{z}}_t = \frac{1}{L} \sum_{l=1}^{L} \hat{\mathbf{z}}_t^{(l)},
\end{equation}
where $\bar{\mathbf{z}}_g \in \mathbb{R}^{d_g}$ and $\bar{\mathbf{z}}_t \in \mathbb{R}^{d_t}$ denote the aggregated graph and text representations, respectively. These aggregated representations are then projected into the unified prototype space using two modality-specific linear transformations:
\begin{equation}
\bar{\mathbf{z}}_{g \rightarrow p} = \mathbf{W}_{g \rightarrow p} \bar{\mathbf{z}}_g + \mathbf{b}_g, \quad
\bar{\mathbf{z}}_{t \rightarrow p} = \mathbf{W}_{t \rightarrow p} \bar{\mathbf{z}}_t + \mathbf{b}_t,
\end{equation}
where $\bar{\mathbf{z}}_{g \rightarrow p}, \bar{\mathbf{z}}_{t \rightarrow p} \in \mathbb{R}^{d_p}$ are the graph and text representations in the unified prototype space, and $\mathbf{W}_{g \rightarrow p} \in \mathbb{R}^{d_g \times d_p}, \mathbf{W}_{t \rightarrow p} \in \mathbb{R}^{d_t \times d_p}$ are learnable projection matrices with modality-specific bias terms $\mathbf{b}_g \in \mathbb{R}^{d_p}$ and $\mathbf{b}_t \in \mathbb{R}^{d_p}$.

To promote prototype selectivity and semantic alignment, we compute the squared Euclidean distance between each projected representation and all prototypes. By flattening the prototype set $\mathcal{P} \in \mathbb{R}^{C \times N \times d_p}$ into a matrix of size $\mathbb{R}^{(C \cdot N) \times d_p}$, we obtain the following distances:
\begin{align}
\mathbf{D}_g & = \left[ \left\| \bar{\mathbf{z}}_{g \rightarrow p} - \mathbf{p}_{c,n} \right\|_2^2 \right]_{(c,n)=1}^{C,N}, \\
\mathbf{D}_t & = \left[ \left\| \bar{\mathbf{z}}_{t \rightarrow p} - \mathbf{p}_{c,n} \right\|_2^2 \right]_{(c,n)=1}^{C,N}, 
\end{align} 
where $\mathbf{D}_g, \mathbf{D}_t \in \mathbb{R}^{C \cdot N}$ represent the prototype-wise squared distances for the graph and text modalities, respectively. For differentiability and numerical stability, these distances are converted to similarity scores using a log-ratio transformation:
\begin{equation}
\mathbf{s}_g = \log\left(\frac{\mathbf{D}_g + 1}{\mathbf{D}_g + \epsilon}\right), \quad
\mathbf{s}_t = \log\left(\frac{\mathbf{D}_t + 1}{\mathbf{D}_t + \epsilon}\right),
\end{equation}
where \( \epsilon \) is a small constant to prevent division by zero.

To enhance discriminability, we retain only the top-$K$ most responsive prototype dimensions through a sparsification step:
\begin{equation}
\hat{\mathbf{s}}_g = \text{TopK}(\mathbf{s}_g, K), \quad
\hat{\mathbf{s}}_t = \text{TopK}(\mathbf{s}_t, K),
\end{equation}
where $\hat{\mathbf{s}}_g = \{ \hat{s}_g^{(1)}, \dots, \hat{s}_g^{(K)} \}$ and $\hat{\mathbf{s}}_t = \{ \hat{s}_t^{(1)}, \dots, \hat{s}_t^{(K)} \}$ are the $K$-dimensional sparse vectors for the graph and text modalities, respectively, with all non-top-$K$ entries masked to zero. These sparse vectors are then normalized into distributions over the selected prototypes using the softmax function:
\begin{equation}
\alpha_{g}^{(k)} = \frac{\exp(\hat{s}_g^{(k)})}{\sum_{j=1}^{K} \exp(\hat{s}_g^{(j)})}, \quad
\alpha_{t}^{(k)} = \frac{\exp(\hat{s}_t^{(k)})}{\sum_{j=1}^{K} \exp(\hat{s}_t^{(j)})},
\end{equation}
where $\boldsymbol{\alpha}_g = \{ \alpha_g^{(1)}, \dots, \alpha_g^{(K)} \} \in \mathbb{R}^K$ and $\boldsymbol{\alpha}_t = \{ \alpha_t^{(1)}, \dots, \alpha_t^{(K)} \} \in \mathbb{R}^K$ represent the prototype distribution for the graph and text modalities, respectively. To promote semantic alignment across modalities, we minimize the Kullback–Leibler divergence between their prototype distributions:
\begin{equation}
\mathcal{L}_{\text{align}} = \mathrm{KL}(\boldsymbol{\alpha}_g \parallel \boldsymbol{\alpha}_t).
\end{equation}

This prototype-level alignment objective encourages both modalities to activate the same subset of semantic prototypes, thereby promoting consistent and discriminative representations within the unified prototype embedding space.

\subsection{Learning Objective and Optimization}

To optimize the proposed framework for accurate molecular property prediction, we formulate a comprehensive training objective. In addition to the prototype alignment loss $\mathcal{L}_{\text{align}}$, we introduce two complementary components to further enhance model performance: (1) a supervised loss $\mathcal{L}_{\text{pred}}$ for downstream classification or regression tasks, and (2) a prototype contrastive loss $\mathcal{L}_{\text{proto}}$, which encourages semantic separation between classes and compactness within each class in the prototype space.

\subsubsection{Task-Specific Predictive Loss}


{To enable task-specific supervision, we employ a predictive loss tailored to the type of downstream task. At the final prediction stage, the model leverages the modality-specific graph representations $\hat{\mathbf{z}}_g^{(i)} \in \mathbb{R}^{d_g}$, which incorporates semantic information transferred from the textual modality, and maps it into the output space through a layer-specific linear transformation:
\begin{equation}
\mathbf{o}^{(i)} = \mathrm{Linear}(\hat{\mathbf{z}}_g^{(i)}).
\end{equation}
The outputs from all hierarchical layers are then aggregated to produce the final prediction:
\begin{equation}
\hat{\mathbf{o}} = \frac{1}{L} \sum_{i=1}^{L} \mathbf{o}^{(i)},
\end{equation}
where $\mathbf{o}^{(i)} \in \mathbb{R}^{C}$ for classification tasks (with $C$ denoting the number of classes), or $\mathbf{o}^{(i)} \in \mathbb{R}$ for regression tasks. This design ensures that the final prediction $\hat{\mathbf{o}}$ captures both the structural dependencies modeled by the GNN and the semantic context transferred from the textual modality through hierarchical bidirectional attention, enabling a unified multimodal representation for molecular property prediction.} 

(1) For classification tasks, the loss $\mathcal{L}^{cl}_{\text{pred}}$ is computed as the cross-entropy between the predicted distribution and the ground-truth label $y \in \{1, \dots, C\}$:
\begin{equation}
\mathcal{L}^{cl}_{\text{pred}} = - \sum_{c=1}^{C} \mathbb{I}(y = c) \log \left( \frac{\exp(\hat{\mathbf{o}}_c)}{\sum_{j=1}^{C} \exp(\hat{\mathbf{o}}_j)} \right),
\end{equation}
where $\hat{\mathbf{o}}_c$ is the aggregated logit for class $c$ and $\mathbb{I}(\cdot)$ is the indicator function. 

(2) For regression tasks, the final prediction $\hat{y} = \hat{\mathbf{o}} \in \mathbb{R}$ is treated as a continuous scalar output, and we employ the mean squared error (MSE) loss to minimize the difference between the predicted and true molecular property values $y$:
\begin{equation}
\mathcal{L}^{re}_{\text{pred}} = \left\| \hat{y} - y \right\|_2^2,
\end{equation}
which directly penalizes deviations from the ground-truth value.

\subsubsection{Prototype Contrastive Loss}

To further enhance the discriminative capacity of the prototype space, we introduce a prototype contrastive loss that encourages modality-specific representations to align closely with semantically relevant prototypes while remaining dissimilar to those associated with other classes. 

(1) For classification tasks, given the unified prototype space $\mathcal{P} = \{ \mathbf{p}_{c, n}\}_{c=1, n=1}^{C, N}$, where $\mathbf{p}_{c, n} \in \mathbb{R}^{d_p}$ denotes the $n$-th prototype of class $c$, each prototype is treated in turn as an anchor. For an anchor prototype $\mathbf{p}_{c, n}$, the remaining prototypes of the same class $\mathcal{P}_c \setminus {\mathbf{p}_{c, n}}$ are considered positive samples, while all prototypes belonging to different classes serve as negatives. The prototype-level contrastive loss for $\mathbf{p}_{c, n}$ is formulated as
\begin{equation}
\mathcal{L}_{\text{proto}}^{c,n} = - \log
\frac{
\frac{1}{N-1} \sum_{\substack{n' = 1, \ n' \neq n}}^{N} \exp \left( \frac{\mathrm{sim}(\mathbf{p}_{c, n}, \mathbf{p}_{c, n'})}{\tau} \right)
}{
\sum_{j=1}^{C} \sum_{m=1}^{N} \exp \left( \frac{\mathrm{sim}(\mathbf{p}_{c, n}, \mathbf{p}_{j, m})}{\tau} \right)
},
\end{equation}
where $\mathrm{sim}(\cdot, \cdot)$ denotes cosine similarity and $\tau$ is a temperature hyperparameter. The overall prototype contrastive loss is then defined as the mean over all prototypes:
\begin{equation}
\mathcal{L}_{\text{proto}}^{cl} = \frac{1}{C N} \sum_{c=1}^{C} \sum_{n=1}^{N} \mathcal{L}_{\text{proto}}^{(c, n)}.
\end{equation}

(2) For regression tasks, the objective is to predict a continuous value whose magnitude reflects the degree or intensity of the molecular property under consideration. For example, in the ESOL dataset, the goal is to predict aqueous solubility, where higher values correspond to greater solubility. To introduce semantic structure into the prototype space and prevent prototype collapse, we employ a two-step approach. First, we apply the K-means algorithm~\cite{hartigan1979algorithm} to partition the prototypes $\mathcal{P}$ into two clusters, denoted as $\mathcal{P}_1$ and $\mathcal{P}_2$, which serve as pseudo-classes reflecting different ranges of the target property. Then, we perform prototype-level contrastive learning within and between these clusters: prototypes belonging to the same cluster are treated as positives, while those from the other cluster are treated as negatives. The prototype contrastive loss is defined as
\begin{equation}
\mathcal{L}_{\text{proto}}^{(m, n)} = - \log
\frac{
\frac{1}{|\mathcal{P}_m| - 1} \sum_{\substack{n' = 1, \ n' \neq n}}^{|\mathcal{P}_m|} \exp\left( \frac{\mathrm{sim}(\mathbf{p}_{m, n}, \mathbf{p}_{m, n'})}{\tau} \right)
}{
\sum_{j=1}^{M} \sum_{l=1}^{|\mathcal{P}_j|} \exp\left( \frac{\mathrm{sim}(\mathbf{p}_{m, n}, \mathbf{p}_{j, l})}{\tau} \right)
},
\end{equation}
where $\mathrm{sim}(\cdot, \cdot)$ denotes cosine similarity, $\tau$ is a temperature hyperparameter, and $|\mathcal{P}_m|$ is the number of prototypes in cluster $m$.

The overall prototype contrastive loss is then defined as the mean over all prototypes:
\begin{equation}
\mathcal{L}_{\text{proto}}^{\text{re}} = \frac{1}{|\mathcal{P}|} \sum_{m=1}^{M} \sum_{n=1}^{|\mathcal{P}_m|} \mathcal{L}_{\text{proto}}^{(m, n)},
\end{equation}
where $|\mathcal{P}|$ denotes the total number of prototypes.

\subsubsection{Unified Training Objective}

The overall training objective is formulated as a weighted combination of all loss components:
\begin{equation}
\mathcal{L}_{\text{total}} = \lambda_{\text{align}} \mathcal{L}_{\text{align}} + \lambda_{\text{pred}} \mathcal{L}_{\text{pred}} + \lambda_{\text{proto}} \mathcal{L}_{\text{proto}},
\end{equation}
where $\lambda_{\text{align}}$, $\lambda_{\text{pred}}$, and $\lambda_{\text{proto}}$ are tunable hyperparameters that balance the contributions of semantic alignment, task-specific supervision, and prototype-level contrastive learning, respectively. This unified loss framework enables end-to-end optimization of cross-modal consistency, predictive accuracy, and representation discriminability in a cohesive manner. {The overall pipeline of \method{} is illustrated in Algorithm~\ref{alg:protomoL}.}

\subsection{Complexity Analysis}

The overall time complexity of \method{} is determined by the dominant components within each module. The dual-branch encoders contribute a complexity of $\mathcal{O}(L \cdot m d_g^2)$ for the $L$-layer GNNs and $\mathcal{O}(L \cdot n^2 d_t)$ for the $L$-layer Transformer, where $m$ and $n$ denote the number of atoms and tokens, respectively. The structured cross-modal interaction module introduces an additional complexity of $\mathcal{O}(L \cdot d^2)$ due to the layer-wise bidirectional attention mechanism. The prototype alignment module incurs a complexity of $\mathcal{O}(C \cdot d_p)$, where $C$ is the number of prototypes. Therefore, the overall time complexity of the proposed \method{} framework is $\mathcal{O}(L(m d_g^2 + n^2 d_t + d^2) + C d_p)$.
\section{Experiments}

\subsection{Experimental Settings}

\begin{table}[t]
    \centering
    \caption{Statistics of the datasets.}
    \vspace{0.5cm}
    \resizebox{0.7\linewidth}{!}{
    \begin{tabular}{l|c|c|c|c|c}
        \hline
        \textbf{Tasks} & \textbf{Dataset} & \textbf{Classes} & \textbf{Graphs} & \textbf{Avg. Nodes} & \textbf{Avg. Edges} \\
        \hline
        \multirow{8}{*}{\textbf{Classification}}
        & HIV & 2 & 41,127 & 25.5 & 27.5 \\
        & BACE & 2 & 1,513 & 34.1 & 36.9 \\
        & BBBP & 2 & 2,039 & 23.9 & 25.2 \\
        & Tox21 & 12 & 7,831 & 18.6 & 19.8 \\
        & ToxCast & 617 & 8,575 & 18.7 & 19.4 \\
        & SIDER & 27 & 1,427 & 33.6 & 35.1 \\
        & ClinTox & 2 & 1,478 & 26.1 & 27.7 \\
        & MUV & 17 & 93,087 & 24.2 & 26.0 \\
        \hline
        \multirow{3}{*}{\textbf{Regression}}
        & Lipo & – & 4,200 & 27.0 & 28.5 \\
        & ESOL & – & 1,128 & 13.3 & 13.9 \\
        & FreeSolv & – & 642 & 8.7 & 9.1 \\
        \hline
    \end{tabular}
    }
    \label{tab:simplified_datasets}
\end{table}

\textbf{Datasets.} To evaluate the effectiveness of the proposed \method{}, we conduct experiments on extensive benchmarks ~\cite{benchmark}. We categorized them into two task types: (1) For classification, we utilize eight molecular property prediction datasets, including ACE, BBBP, HIV, ClinTox, Tox21, MUV, SIDER, and ToxCast. These datasets cover a diverse range of biochemical properties, such as blood–brain barrier permeability (BBBP), molecular toxicity (ClinTox, Tox21), and biological activity (HIV, MUV), spanning both binary and multi-label classification settings; (2) For regression, we adopt three datasets: ESOL, FreeSolv, and Lipophilicity, which involve predicting continuous-valued molecular properties including aqueous solubility (ESOL), hydration free energy (FreeSolv), and lipophilicity (Lipo). These tasks present unique challenges due to the fine-grained structural and semantic variations across molecules. The statistics of the above datasets is provided in Table \ref{tab:simplified_datasets}. {Additionally, we adopt a standard random stratified split of the dataset into training (80\%), testing (10\%), and validation (10\%) sets.}

{\textbf{Baselines.} We compare the proposed \method{} with various competitive baselines across the evaluated datasets, including two textual methods: SMILES2vec~\cite{goh2017smiles2vec} and SMILES-BERT~\cite{wang2019smiles}; fourteen graph-based methods: GraphPT~\cite{CandA}, GraphSAGE~\cite{EdgePred}, DGI~\cite{Infomax}, JOAOv2~\cite{JOAO}, GraphCL~\cite{GraphCL}, GraphLoG~\cite{graphlog}, MICRO-Graph~\cite{micrograph}, MGSSL~\cite{mgssl}, GraphFP~\cite{graphft}, GROVE~\cite{grove}, SimSGT~\cite{simsgt}, MoAMa~\cite{moama}, Uni-Mol~\cite{Uni-Mol} and S-CGIB~\cite{S-CGIB}; and three multimodal methods that integrate textual and graph-based representations: Tri-SGD~\cite{lu2024multimodal}, MMSG~\cite{wu2023molecular}, and MDFCL~\cite{gong2025mdfcl}. More details about baselines are introduced in Appendix \ref{sec:baselines}.

\textbf{Evaluation Metrics.}
Following the prior work~\cite{S-CGIB}, we adopt two widely recognized metrics tailored to the nature of the prediction task: the Area Under the Receiver Operating Characteristic Curve (ROC-AUC) for classification, and the Root Mean Squared Error (RMSE) for regression. 

\begin{itemize}
\item \textbf{ROC-AUC.} For classification tasks, we report the Area Under the Receiver Operating Characteristic Curve (ROC-AUC). The ROC-AUC quantifies the model's ability to distinguish between classes by integrating the trade-off between the True Positive Rate (TPR) and the False Positive Rate (FPR) across all possible thresholds. The TPR and FPR are defined as:
\begin{equation*}
\mathrm{TPR} = \frac{\mathrm{True\ Positive}}{\mathrm{True\ Positive} + \mathrm{False\ Negative}},
\end{equation*}
\begin{equation*}
\mathrm{FPR} = \frac{\mathrm{False\ Positive}}{\mathrm{False\ Positive} + \mathrm{True\ Negative}}.
\end{equation*}
A higher ROC-AUC value indicates superior classification performance, reflecting a stronger capacity to discriminate between positive and negative classes.

\begin{table*}[t] 
\centering
\caption{Classification results (ROC-AUC \%) for molecular property prediction on the BACE, BBBP, HIV, ClinTox, Tox21, MUV, SIDER, and ToxCast datasets. \textbf{Bold} results indicate the best performance.} 
\resizebox{\linewidth}{!}{
\begin{tabular}{l|c|c|c|c|c|c|c|c}
\hline
\textbf{Methods} & \textbf{BACE} & \textbf{BBBP} & \textbf{HIV} & \textbf{ClinTox} & \textbf{Tox21} & \textbf{MUV} & \textbf{SIDER} & \textbf{ToxCast} \\
\hline
SMILES2vec  & 81.4\scriptsize{$\pm$3.6} & 78.0\scriptsize{$\pm$2.4} & 75.8\scriptsize{$\pm$3.2} & 74.1\scriptsize{$\pm$2.2} & 73.4\scriptsize{$\pm$2.0} & 72.3\scriptsize{$\pm$3.4} & 56.7\scriptsize{$\pm$1.9} & 62.2\scriptsize{$\pm$1.9} \\
SMILES-BERT & 83.3\scriptsize{$\pm$3.3} & 80.5\scriptsize{$\pm$2.2} & 77.9\scriptsize{$\pm$2.9} & 76.2\scriptsize{$\pm$2.1} & 75.1\scriptsize{$\pm$1.9} & 73.7\scriptsize{$\pm$3.2} & 58.0\scriptsize{$\pm$1.8} & 64.5\scriptsize{$\pm$1.8} \\
\hline
GraphPT & 76.0\scriptsize{$\pm$0.5} & 67.1\scriptsize{$\pm$0.5} & 72.7\scriptsize{$\pm$0.7} & 60.1\scriptsize{$\pm$1.2} & 73.4\scriptsize{$\pm$0.6} & 67.9\scriptsize{$\pm$0.6} & 61.2\scriptsize{$\pm$0.7} & 61.7\scriptsize{$\pm$1.2} \\
GraphSAGE & 74.3\scriptsize{$\pm$1.4} & 64.7\scriptsize{$\pm$1.1} & 70.6\scriptsize{$\pm$1.7} & 61.6\scriptsize{$\pm$1.3} & 70.3\scriptsize{$\pm$1.6} & 70.8\scriptsize{$\pm$1.6} & 60.2\scriptsize{$\pm$0.8} & 60.0\scriptsize{$\pm$0.8} \\
DGI & 77.8\scriptsize{$\pm$0.5} & 68.4\scriptsize{$\pm$0.6} & 73.6\scriptsize{$\pm$0.5} & 58.6\scriptsize{$\pm$0.8} & 72.7\scriptsize{$\pm$0.2} & 72.1\scriptsize{$\pm$1.2} & 59.0\scriptsize{$\pm$0.6} & 62.8\scriptsize{$\pm$0.5} \\
JOAOv2 & 74.4\scriptsize{$\pm$1.7} & 72.0\scriptsize{$\pm$0.2} & 77.1\scriptsize{$\pm$1.5} & 65.2\scriptsize{$\pm$0.8} & 74.0\scriptsize{$\pm$1.9} & 68.5\scriptsize{$\pm$1.6} & 59.9\scriptsize{$\pm$1.7} & 63.1\scriptsize{$\pm$1.9} \\
GraphCL & 77.8\scriptsize{$\pm$0.5} & 68.4\scriptsize{$\pm$0.6} & 73.6\scriptsize{$\pm$0.5} & 61.6\scriptsize{$\pm$1.3} & 73.3\scriptsize{$\pm$0.6} & 72.1\scriptsize{$\pm$1.2} & 61.8\scriptsize{$\pm$0.6} & 62.8\scriptsize{$\pm$0.5} \\
GraphLoG & 76.6\scriptsize{$\pm$1.0} & 66.8\scriptsize{$\pm$0.3} & 73.8\scriptsize{$\pm$0.3} & 53.8\scriptsize{$\pm$0.9} & 71.6\scriptsize{$\pm$0.5} & 72.5\scriptsize{$\pm$2.0} & 59.1\scriptsize{$\pm$0.5} & 61.5\scriptsize{$\pm$0.4} \\
GraphFP & 80.3\scriptsize{$\pm$3.1} & 72.1\scriptsize{$\pm$1.2} & 75.7\scriptsize{$\pm$1.4} & 76.8\scriptsize{$\pm$1.8} & 77.4\scriptsize{$\pm$1.4} & 71.8\scriptsize{$\pm$1.3} & 65.9\scriptsize{$\pm$3.1} & 69.2\scriptsize{$\pm$1.9} \\
MICRO-Graph & 63.6\scriptsize{$\pm$1.6} & 67.2\scriptsize{$\pm$1.9} & 76.7\scriptsize{$\pm$1.1} & 77.6\scriptsize{$\pm$1.6} & 71.8\scriptsize{$\pm$1.7} & 70.5\scriptsize{$\pm$1.6} & 60.3\scriptsize{$\pm$1.0} & 60.8\scriptsize{$\pm$1.2} \\
MGSSL & 82.0\scriptsize{$\pm$3.8} & 79.5\scriptsize{$\pm$2.0} & 77.5\scriptsize{$\pm$2.9} & 75.8\scriptsize{$\pm$1.8} & 74.8\scriptsize{$\pm$1.6} & 73.4\scriptsize{$\pm$3.5} & 57.5\scriptsize{$\pm$1.5} & 63.9\scriptsize{$\pm$1.6} \\
GROVE & 81.1\scriptsize{$\pm$0.1} & 87.1\scriptsize{$\pm$0.1} & 75.0\scriptsize{$\pm$0.1} & 72.5\scriptsize{$\pm$0.1} & 68.6\scriptsize{$\pm$0.2} & 67.7\scriptsize{$\pm$0.1} & 57.5\scriptsize{$\pm$0.2} & 64.4\scriptsize{$\pm$0.1} \\
SimSGT & 79.8\scriptsize{$\pm$1.3} & 71.5\scriptsize{$\pm$1.8} & 78.1\scriptsize{$\pm$1.1} & 74.1\scriptsize{$\pm$1.1} & 76.2\scriptsize{$\pm$1.3} & 72.8\scriptsize{$\pm$1.5} & 59.7\scriptsize{$\pm$1.3} & 65.8\scriptsize{$\pm$0.8} \\
MoAMa & 81.3\scriptsize{$\pm$1.1} & 85.9\scriptsize{$\pm$0.6} & 78.1\scriptsize{$\pm$0.6} & 77.1\scriptsize{$\pm$1.7} & 78.3\scriptsize{$\pm$0.6} & 72.4\scriptsize{$\pm$1.8} & 62.7\scriptsize{$\pm$0.4} & 68.0\scriptsize{$\pm$1.1} \\
Uni-Mol & 85.7\scriptsize{$\pm$0.2} & 72.9\scriptsize{$\pm$0.6} & 80.8\scriptsize{$\pm$0.3} & 91.9\scriptsize{$\pm$1.8} & 79.6\scriptsize{$\pm$0.5} & \textbf{82.1\scriptsize{$\pm$1.3}} & 65.9\scriptsize{$\pm$1.3} & 69.6\scriptsize{$\pm$0.1} \\
S-CGIB & 86.5\scriptsize{$\pm$0.8} & 88.8\scriptsize{$\pm$0.5} & 78.3\scriptsize{$\pm$1.3} & 78.6\scriptsize{$\pm$2.0} & 80.9\scriptsize{$\pm$0.2} & 77.7\scriptsize{$\pm$1.2} & 64.0\scriptsize{$\pm$1.0} & 71.0\scriptsize{$\pm$0.3} \\
\hline
Tri\_SGD & 90.3\scriptsize{$\pm$1.0} & 89.4\scriptsize{$\pm$0.5} & 80.9\scriptsize{$\pm$0.9} & 83.0\scriptsize{$\pm$0.8} & 80.7\scriptsize{$\pm$0.6} & 79.4\scriptsize{$\pm$1.0} & 67.8\scriptsize{$\pm$0.8} & 70.9\scriptsize{$\pm$0.9} \\
MMSG & 90.8\scriptsize{$\pm$0.5} & 89.7\scriptsize{$\pm$0.2} & 81.0\scriptsize{$\pm$0.3} & 85.0\scriptsize{$\pm$0.7} & 81.0\scriptsize{$\pm$1.0} & 79.3\scriptsize{$\pm$0.7} & 67.8\scriptsize{$\pm$0.5} & 71.0\scriptsize{$\pm$1.4} \\
MDFCL & 86.4\scriptsize{$\pm$1.0} & 78.4\scriptsize{$\pm$1.0} & 78.7\scriptsize{$\pm$1.3} & \textbf{93.7\scriptsize{$\pm$2.0}} & 80.5\scriptsize{$\pm$0.7} & 78.5\scriptsize{$\pm$0.7} & 67.7\scriptsize{$\pm$0.6} & 70.9\scriptsize{$\pm$0.8} \\
\hline
\method{} & \textbf{91.4\scriptsize{$\pm$0.3}} & \textbf{90.3\scriptsize{$\pm$0.6}} & \textbf{81.2\scriptsize{$\pm$0.2}} & 84.3\scriptsize{$\pm$0.3} & \textbf{81.2\scriptsize{$\pm$0.3}} & 80.7\scriptsize{$\pm$0.2} & \textbf{68.1\scriptsize{$\pm$0.5}} & \textbf{71.2\scriptsize{$\pm$0.4}} \\
\hline
\end{tabular}
}
\label{tab:main_classificaition}
\end{table*}

\item \textbf{RMSE.} For regression tasks, we utilize the Root Mean Squared Error (RMSE) to measure the accuracy of predicted continuous-valued molecular properties. RMSE is computed as:
\begin{equation*}
    \mathrm{RMSE} = \sqrt{ \frac{1}{N} \sum_{i=1}^N (y_i - \hat{y}_i)^2 },
\end{equation*}
where $y_i$ and $\hat{y}_i$ denote the ground-truth and predicted values for the $i$-th molecule, and $N$ is the total number of samples. Lower RMSE values indicate higher regression accuracy, signifying better agreement between predicted and actual property values.

\end{itemize}

\begin{figure*}[t]
    \centering
    \includegraphics[width=1.0\linewidth]{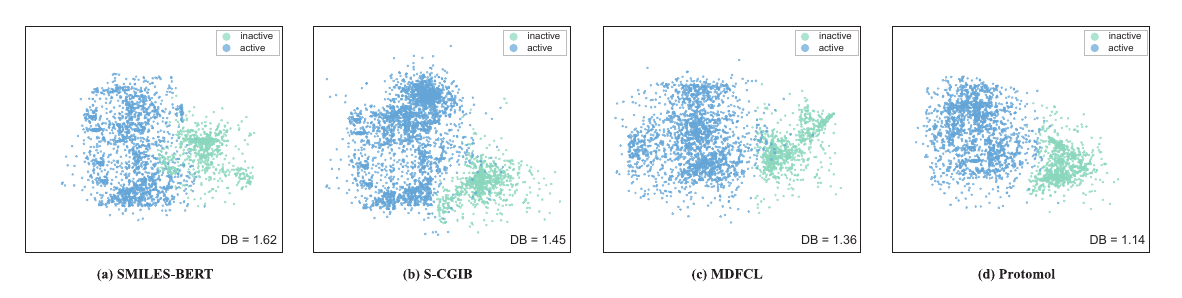} 
    \caption{Visualization results of baseline methods and \method{} on the HIV dataset.}
    \label{fig:vis_hiv}
\end{figure*}

\begin{figure*}[t]
    \centering\includegraphics[width=1.0\linewidth]{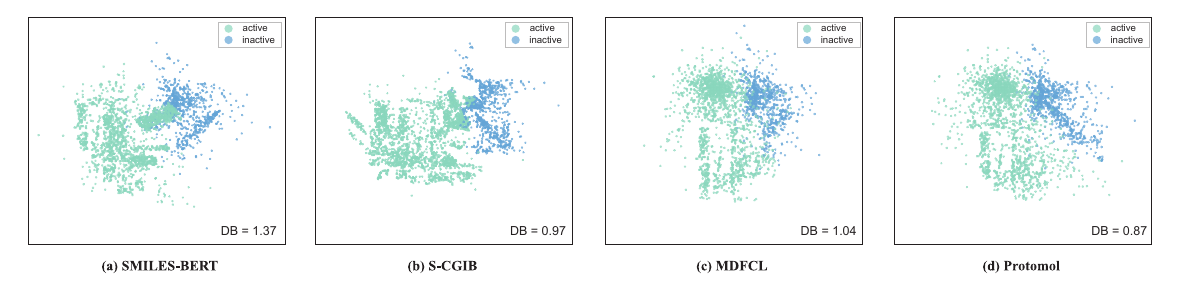} 
    \caption{Visualization results of baseline methods and \method{} on the BBBP dataset.}
    \label{fig:vis_bbbp}
\end{figure*}

\textbf{Implementation Details.} We implement the proposed \method{} using the PyTorch framework~\footnote{\url{https://pytorch.org/}}, and re-run all baselines within the same framework to ensure a fair comparison. All experiments are conducted on the NVIDIA RTX 4090 GPUs. We employ the Adam optimizer with a learning rate of $8 \times 10^{-5}$ and a weight decay of $1 \times 10^{-4}$. A cosine annealing learning rate schedule is used, with the maximum number of epochs set to 100. The proposed \method{} and all baselines are trained for a total of 100 epochs with a batch size of 128. Additionally, for \method{}, we select GIN~\cite{xu2018powerful} as the backbone for the graph encoder to effectively capture complex molecular structures, and Qwen-2.5 7B~\cite{yang2025qwen3} as the backbone for the textual encoder to extract rich semantic information from textual representations. All reported results are averaged over five independent runs.

\subsection{Performance Comparison}

We present the results of the proposed \method{} with all baseline models on two types of molecular property prediction tasks: classification and regression in Table~\ref{tab:main_classificaition} and Table~\ref{tab:main_regression}. 

\begin{wraptable}{r}{0.55\linewidth} 
\vspace{-0.4cm}
\caption{Regression results (RMSE) for molecular property prediction on the ESOL, FreeSolv, and Lipophilicity datasets. \textbf{Bold} results indicate the best performance.}
\vspace{-0.2cm}
\centering
\resizebox{0.99\linewidth}{!}{ 
\begin{tabular}{l|c|c|c}
\hline
\textbf{Methods} & \textbf{ESOL} & \textbf{FreeSolv} & \textbf{Lipophilicity} \\
\hline
SMILES2vec  & 2.802\scriptsize{$\pm$0.108} & 2.978\scriptsize{$\pm$0.372} & 1.068\scriptsize{$\pm$0.086} \\
SMILES-BERT & 2.612\scriptsize{$\pm$0.085} & 2.354\scriptsize{$\pm$0.061} & 0.832\scriptsize{$\pm$0.079} \\
\hline
GraphPT & 2.954\scriptsize{$\pm$0.087} & 4.023\scriptsize{$\pm$0.039} & 0.982\scriptsize{$\pm$0.052} \\
GraphSAGE & 2.368\scriptsize{$\pm$0.070} & 3.192\scriptsize{$\pm$0.023} & 1.085\scriptsize{$\pm$0.061} \\
DGI & 2.953\scriptsize{$\pm$0.049} & 3.033\scriptsize{$\pm$0.026} & 0.970\scriptsize{$\pm$0.023} \\
JOAOv2 & 2.144\scriptsize{$\pm$0.009} & 3.842\scriptsize{$\pm$0.012} & 1.116\scriptsize{$\pm$0.024} \\
GraphCL & 1.390\scriptsize{$\pm$0.363} & 3.166\scriptsize{$\pm$0.027} & 1.014\scriptsize{$\pm$0.018} \\
GraphLoG & 1.542\scriptsize{$\pm$0.026} & 2.335\scriptsize{$\pm$0.052} & 0.932\scriptsize{$\pm$0.052} \\
GraphFP & 2.136\scriptsize{$\pm$0.096} & 2.528\scriptsize{$\pm$0.016} & 1.371\scriptsize{$\pm$0.058} \\
MICRO-Graph & 0.842\scriptsize{$\pm$0.055} & 1.865\scriptsize{$\pm$0.061} & 0.851\scriptsize{$\pm$0.073} \\
MGSSL & 2.936\scriptsize{$\pm$0.071} & 2.940\scriptsize{$\pm$0.051} & 1.106\scriptsize{$\pm$0.077} \\
GROVE & 1.237\scriptsize{$\pm$0.403} & 2.712\scriptsize{$\pm$0.327} & 0.823\scriptsize{$\pm$0.027} \\
SimSGT & 0.932\scriptsize{$\pm$0.026} & 1.953\scriptsize{$\pm$0.038} & 0.771\scriptsize{$\pm$0.041} \\
MoAMa & 1.125\scriptsize{$\pm$0.029} & 2.072\scriptsize{$\pm$0.053} & 1.085\scriptsize{$\pm$0.024} \\
Uni-Mol & 0.788\scriptsize{$\pm$0.029} & 1.620\scriptsize{$\pm$0.035} & 0.603\scriptsize{$\pm$0.010} \\
S-CGIB & 0.816\scriptsize{$\pm$0.019} & 1.648\scriptsize{$\pm$0.074} & 0.762\scriptsize{$\pm$0.042} \\
\hline
Tri\_SGD & 0.654\scriptsize{$\pm$0.106} & 1.607\scriptsize{$\pm$0.114} & 0.606\scriptsize{$\pm$0.023} \\
MMSG & 0.688\scriptsize{$\pm$0.030} & 1.638\scriptsize{$\pm$0.525} & \textbf{0.577\scriptsize{$\pm$0.014}} \\
MDFCL & 0.663\scriptsize{$\pm$0.026} & 1.607\scriptsize{$\pm$0.032} & 0.608\scriptsize{$\pm$0.037} \\
\hline
\method{} & \textbf{0.629\scriptsize{$\pm$0.014}} & \textbf{1.522\scriptsize{$\pm$0.044}} & 0.583\scriptsize{$\pm$0.039} \\
\hline
\end{tabular}
}
\label{tab:main_regression}
\end{wraptable}

From these tables, we observe that: (1) Graph-based methods, such as S-CGIB~\cite{S-CGIB} and Uni-Mol~\cite{Uni-Mol}, consistently outperform text-based approaches by more effectively capturing intrinsic molecular topology and relational structure. While text-based models like SMILES2vec and SMILES-BERT encode molecules as linear sequences, they are inherently limited in their ability to represent the complex connectivity patterns, stereochemistry, and spatial dependencies present in molecular graphs. In contrast, graph-based approaches operate directly on molecular graphs and leverage message passing mechanisms to aggregate both local and global structural information, which enables the construction of richer and more chemically meaningful representations for molecules. (2) Multimodal methods, such as MDFCL~\cite{gong2025mdfcl} and MMSG~\cite{wu2023molecular}, which integrate both graph and textual information, consistently surpass graph-based models by harnessing the complementary strengths of each modality. These approaches combine the structural inductive bias of Graph Neural Networks (GNNs) with the semantic richness provided by textual encoders, allowing the models to jointly reason over atomic connectivity, subgraph motifs, and higher-level chemical semantics present in SMILES or other text representations. (3) The proposed \method{} consistently achieves superior performance over all baselines across most tasks, attributable to its architecture that integrates hierarchical cross-modal interaction with prototype-guided semantic alignment. Different from conventional methods that typically perform fusion at the final encoder layer, \method{} introduces a structured, layer-wise bidirectional cross-modal attention mechanism. This design enables progressive and fine-grained semantic integration between graph and textual representations at multiple levels, thereby preserving the distinctive strengths of each modality throughout the representation learning process. Furthermore, \method{} employs a unified semantic prototype space comprising learnable, class-specific anchors that act as shared semantic groundings for both modalities. This space not only facilitates consistent cross-modal alignment but also underpins prototype-guided contrastive learning, which enhances intra-class compactness and inter-class separability in the joint embedding space. More analysis can be found in Appendix.~\ref{sec:vis_new}.

Additionally, we present comparative visualization results for \method{}, SMILES-BERT~\cite{wang2019smiles}, S-CGIB~\cite{S-CGIB}, and MDFCL~\cite{gong2025mdfcl} on the HIV and BBBP datasets, as shown in Figure.~\ref{fig:vis_hiv} and \ref{fig:vis_bbbp}. These results demonstrate that \method{} yields substantially clearer class boundaries and more accurate classification outcomes compared to the baseline methods. {Moreover, a quantitative evaluation using the Davies–Bouldin (DB) Index~\cite{xiao2017davies} shows that \method{} attains lower DB values (1.14 on HIV and 0.87 on BBBP) than all baselines, indicating stronger inter-class separability and more discriminative feature representations.}

\begin{figure}[t]
    \centering
    \begin{subfigure}{0.49\linewidth}
        \centering
        \includegraphics[width=\linewidth]{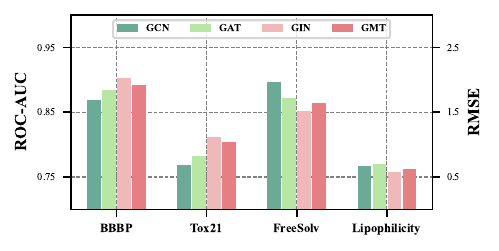}
        \caption{Graph encoder backbones.}
        \label{fig:vis_graph}
    \end{subfigure}\hfill
    \begin{subfigure}{0.49\linewidth}
        \centering
        \includegraphics[width=\linewidth]{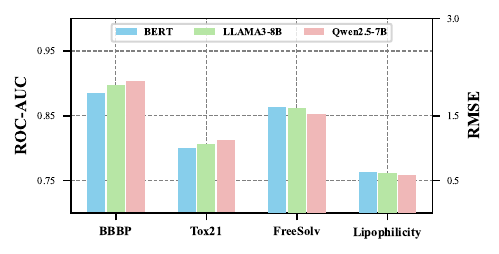}
        \caption{Textual encoder backbones.}
        \label{fig:vis_text}
    \end{subfigure}
    \caption{Performance of different backbones on BBBP and Tox21 (classification) and on FreeSolv and Lipophilicity (regression): 
    (a) graph encoder backbones; (b) textual encoder backbones.}
    \label{fig:backbone_compare}
\end{figure}

\subsection{Flexible Architecture}

To assess the impact of different backbone choices for the graph and textual encoders in \method{}, we systematically evaluate several state-of-the-art architectures. For the graph encoder, we consider a range of message passing methods, including GCN~\cite{kipf2016semi}, GAT~\cite{velickovic2017graph}, GIN~\cite{xu2018powerful}, and GMT~\cite{baek2021accurate}. For the textual encoder, we experiment with pretrained language models such as BERT~\cite{devlin2019bert}, LLaMA3-8B~\cite{grattafiori2024llama}, and Qwen2.5-7B~\cite{yang2025qwen3}.

As illustrated in Figure.~\ref{fig:vis_graph} and \ref{fig:vis_text}, we observe that both GIN and Qwen2.5-7B consistently outperform other backbone settings, underscoring their superior capacity to capture the complex structural patterns and semantic nuances critical for molecular property prediction. These findings further justify our design choice of adopting GIN and Qwen2.5-7B as the core components of \method{}, as they provide a strong foundation for learning robust and discriminative multimodal molecular representations.
\subsection{Ablation Study}

\begin{wraptable}{r}{0.55\linewidth}  
\vspace{-1.2cm}
\caption{The results of ablation studies on the regression task across the ESOL, FreeSolv, and Lipophilicity datasets. \textbf{Bold} results indicate the best performance.}
\vspace{0.2cm}
\centering
\resizebox{\linewidth}{!}{
\begin{tabular}{l|c|c|c}
\hline
\textbf{Methods} & \textbf{ESOL} & \textbf{FreeSolv} & \textbf{Lipophilicity} \\
\hline
\method{} w/o CA & 0.712\scriptsize{$\pm$0.097} & 1.657\scriptsize{$\pm$0.082} & 0.638\scriptsize{$\pm$0.084} \\
\method{} w/o UP & 0.691\scriptsize{$\pm$0.072} & 1.657\scriptsize{$\pm$0.088} & 0.627\scriptsize{$\pm$0.069} \\
\method{} w/o AL & 0.705\scriptsize{$\pm$0.083} & 1.865\scriptsize{$\pm$0.074} & 0.695\scriptsize{$\pm$0.082} \\
\method{} w/o CL & 0.810\scriptsize{$\pm$0.107} & 1.698\scriptsize{$\pm$0.094} & 0.666\scriptsize{$\pm$0.072} \\
\method{} w/o PR & 0.735\scriptsize{$\pm$0.079} & 1.739\scriptsize{$\pm$0.082} & 0.817\scriptsize{$\pm$0.061} \\
\hline
\method{} & \textbf{0.629\scriptsize{$\pm$0.014}} & \textbf{1.522\scriptsize{$\pm$0.044}} & \textbf{0.583\scriptsize{$\pm$0.039}} \\
\hline
\end{tabular}
}
\label{tab:ablation_regression}
\vspace{-0.3cm}
\end{wraptable}

We conduct ablation studies to examine the contributions of each component in the proposed \method{}. The ablations are categorized into two groups: (1) structural components, which govern cross-modal representation learning; and (2) objective components, which determine the optimization behavior of the model.

\begin{table*}[t!] 
\centering
\caption{The results of ablation studies on the classification task across the BACE, BBBP, HIV, ClinTox, Tox21, MUV, SIDER, and ToxCast datasets. \textbf{Bold} results indicate the best performance.} 
\resizebox{1.0\linewidth}{!}{
\begin{tabular}{l|c|c|c|c|c|c|c|c}
\hline
\textbf{Methods} & \textbf{BACE} & \textbf{BBBP} & \textbf{HIV} & \textbf{ClinTox} & \textbf{Tox21} & \textbf{MUV} & \textbf{SIDER} & \textbf{ToxCast} \\
\hline

\method{} w/o CA & 90.1\scriptsize{$\pm$0.4} & 88.7\scriptsize{$\pm$0.6} & 80.7\scriptsize{$\pm$0.6} & 79.6\scriptsize{$\pm$0.4} & 79.5\scriptsize{$\pm$0.6} & 78.1\scriptsize{$\pm$0.7} & 67.2\scriptsize{$\pm$0.7} & 70.4\scriptsize{$\pm$0.8} \\
\method{} w/o UP & 89.4\scriptsize{$\pm$0.3} & 85.6\scriptsize{$\pm$0.7} & 80.1\scriptsize{$\pm$0.6} & 81.4\scriptsize{$\pm$0.6} & 79.3\scriptsize{$\pm$0.4} & 75.2\scriptsize{$\pm$0.9} & 64.7\scriptsize{$\pm$0.8} & 70.2\scriptsize{$\pm$0.7} \\
\method{} w/o AL & 88.5\scriptsize{$\pm$0.5} & 87.9\scriptsize{$\pm$0.6} & 78.4\scriptsize{$\pm$0.8} & 77.9\scriptsize{$\pm$0.8} & 78.1\scriptsize{$\pm$0.7} & 77.0\scriptsize{$\pm$0.8} & 61.0\scriptsize{$\pm$0.7} & 68.7\scriptsize{$\pm$0.7} \\
\method{} w/o CL & 84.3\scriptsize{$\pm$0.7} & 85.3\scriptsize{$\pm$0.7} & 78.9\scriptsize{$\pm$0.4} & 78.2\scriptsize{$\pm$0.5} & 77.3\scriptsize{$\pm$0.5} & 70.3\scriptsize{$\pm$1.2} & 62.1\scriptsize{$\pm$0.8} & 66.0\scriptsize{$\pm$0.7} \\
\method{} w/o PR & 87.3\scriptsize{$\pm$0.5} & 82.9\scriptsize{$\pm$0.7} & 78.1\scriptsize{$\pm$0.6} & 76.9\scriptsize{$\pm$0.5} & 75.7\scriptsize{$\pm$0.4} & 73.1\scriptsize{$\pm$1.1} & 60.2\scriptsize{$\pm$0.9} & 67.1\scriptsize{$\pm$0.5} \\

\hline

\method{} & \textbf{91.4\scriptsize{$\pm$0.3}} & \textbf{90.3\scriptsize{$\pm$0.6}} & \textbf{81.2\scriptsize{$\pm$0.2}} & \textbf{84.3\scriptsize{$\pm$0.3}} & \textbf{81.2\scriptsize{$\pm$0.3}} & \textbf{80.7\scriptsize{$\pm$0.2}} & \textbf{68.1\scriptsize{$\pm$0.5}} & \textbf{71.2\scriptsize{$\pm$0.4}} \\
\hline
\end{tabular}
}
\label{tab:ablation_classificaition}
\vspace{-0.5cm}
\end{table*}

(1) For structural components, we conduct two variants of \method{}: \method{} w/o CA: it removes the hierarchical cross-modal attention and performs cross-modal interaction only at the final layer, and \method{} w/o UP: it removes the unified prototype space and applies separate prototype learning for the graph and text modalities independently. Experimental results for the structural components are reported in Table~\ref{tab:ablation_regression} and ~\ref{tab:ablation_classificaition}. From the results, we find that \method{} outperforms \method{} w/o CA, demonstrating the essential role of hierarchical cross-modal attention. This mechanism enables progressive, layer-wise semantic integration between graph and textual modalities, leading to unified and informative molecular representations that substantially improve predictive accuracy. \method{} also surpasses \method{} w/o UP, highlighting the significance of the unified prototype space. By introducing shared semantic anchors across modalities, this component enhances cross-modal alignment and enables the model to jointly capture complementary structural and semantic information. (2) For objective components, we evaluate three variants of \method{}: \method{} w/o AL, which removes the alignment loss $\mathcal{L}_{\text{align}}$ that enforces semantic coherence between graph and text representations; \method{} w/o CL, which eliminates the prediction loss $\mathcal{L}_{\text{pred}}$ providing task-specific supervision; and \method{} w/o PR, which excludes the prototype-guided contrastive loss $\mathcal{L}_{\text{proto}}$ designed to strengthen inter-class separation and enhance the discriminative structure of the embedding space. As shown in Table~\ref{tab:ablation_classificaition} and \ref{tab:ablation_regression}, \method{} w/o AL, \method{} w/o CL, and \method{} w/o PR all exhibit noticeably degraded performance compared to the full \method{}, confirming the necessity of the alignment, prediction, and prototype-guided contrastive losses. These objectives collectively ensure semantic consistency, provide reliable supervised signals, and promote well-structured embedding spaces, thereby improving the discriminative power and generalization capability of \method{} across a wide range of molecular property prediction tasks.

\begin{figure*}[t]
    \centering
    \includegraphics[width=1.0\linewidth]{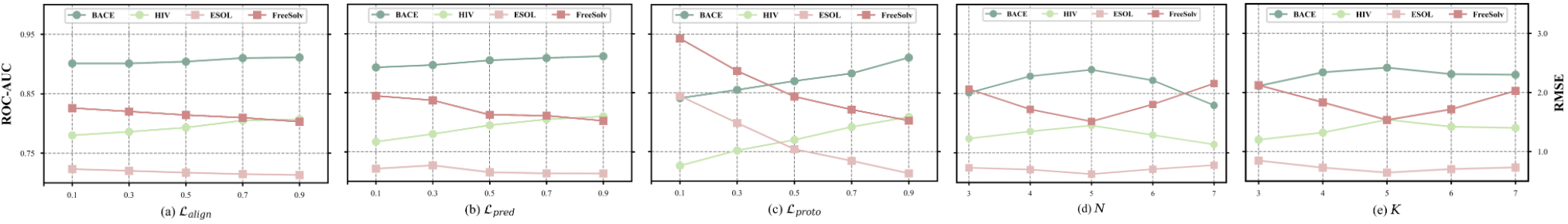} 
    \caption{{Hyperparameter sensitivity analysis of $\lambda_\text{align}$, $\lambda_\text{pred}$, $\lambda_\text{proto}$,
    number of prototype $N$, and number of $K$ entries
    on the BACE and HIV datasets for classification tasks, and on the ESOL and FreeSolv datasets for regression tasks.}}
    \label{fig:hyper}
\end{figure*}

\subsection{Sensitivity Analysis}

We perform a sensitivity analysis to assess how the key hyperparameters of \method{}, including the alignment loss coefficient $\lambda_{\text{align}}$, the prediction loss coefficient $\lambda_{\text{pred}}$, and the prototype-guided contrastive loss coefficient $\lambda_{\text{proto}}$, {as well as the number of prototypes per class $N$ and the number of top-$K$ activated prototype entries, collectively influence its overall performance.} In particular, $\lambda_{\text{align}}$ modulates the strength of semantic alignment between graph and textual representations, $\lambda_{\text{pred}}$ controls the contribution of the task-specific prediction objective, and $\lambda_{\text{proto}}$ determines the impact of prototype-guided contrastive learning on the structure of the joint embedding space. These parameters are critical in balancing supervision quality and model robustness. {Meanwhile, the number of prototypes per class $N$ controls the granularity of semantic abstraction within each category, and the number of top-$K$ activated prototype entries governs how selectively the model attends to representative prototypes during alignment.}

{Figure.~\ref{fig:hyper} illustrates how $\lambda_{\text{align}}$, $\lambda_{\text{pred}}$, $\lambda_{\text{proto}}$, as well as the number of prototypes per class $N$ and the number of top-$K$ activated prototype entries, influence the performance of \method{} on the BACE, HIV, ESOL, and FreeSolv datasets. Each loss coefficient is varied within the range $\{0.1, 0.3, 0.5, 0.7, 0.9\}$, while $N$ and $K$ are varied within $\{3, 4, 5, 6, 7\}$.} From the results, we observe that: (1) The performance of \method{} in Figure.~\ref{fig:hyper}(a) steadily improves as $\lambda_{\text{align}}$ increases. Increasing the strength of semantic alignment between graph and textual modalities enables the model to learn more consistent and robust multimodal representations. This enhanced alignment bridges the modality gap and promotes more effective information integration, ultimately leading to superior predictive performance across a wide range of molecular property prediction tasks. Thus, we set $\lambda_{\text{align}}$ to 0.9 as the default. (2) As shown in Figure.~\ref{fig:hyper}(b), increasing $\lambda_{\text{pred}}$, which governs the contribution of the supervised information, consistently improves performance across both classification and regression tasks, highlighting the importance of balancing supervision strength within the overall objective for effective model optimization. Thus, we set $\lambda_{\text{pred}}$ to 0.9 as the default. (3) Figure.~\ref{fig:hyper}(c) demonstrates that higher values of $\lambda_{\text{proto}}$ further improve performance, as the prototype-guided contrastive loss encourages the formation of well-structured and discriminative embedding spaces. This improved embedding structure facilitates greater intra-class compactness and inter-class separability, enhancing the model's generalization ability across diverse molecular property prediction tasks. Thus, we set $\lambda_{\text{proto}}$ to 0.9 as the default. {(4) Figure.~\ref{fig:hyper}(d) and (e) illustrate that the performance of \method{} initially improves and then gradually declines as the number of prototypes per class $N$ and the number of top-$K$ activated prototype entries increase. Moderate settings ($N=5$, $K=5$) yield the best results, suggesting that an optimal trade-off between semantic diversity and representational sparsity is essential for effective prototype learning. Insufficient values of $N$ or $K$ restrict the expressiveness of the prototype space, whereas excessively large values introduce redundancy and noise, undermining discriminative capability. Therefore, we set $N=5$ and $K=5$ as the default configuration in all experiments.}
\section{Conclusion}

In this work, we introduced \method{}, a prototype-guided cross-modal molecular representation learning framework that enables fine-grained semantic integration between molecular graphs and textual descriptions. By leveraging dual-branch hierarchical encoders and a structured, layer-wise bidirectional cross-modal attention mechanism, \method{} captures complex structural and semantic dependencies at multiple levels. The introduction of a unified semantic prototype space further ensures robust modality-invariant alignment and enhances the discriminative power of molecular embeddings. Extensive experiments across diverse molecular property prediction benchmarks demonstrate that \method{} consistently outperforms state-of-the-art methods in both classification and regression settings, advancing the field of multimodal molecular representation learning. In the future, we plan to extend \method{} to support additional modalities, such as protein structures and biological pathways, and to explore its application in large-scale, real-world bioinformatics scenarios including drug-target interaction prediction and functional annotation.

\bibliographystyle{plain}
\bibliography{main}
\onecolumn
\appendix
\section{Appendix}

\subsection{More details about baseline methods}\label{sec:baselines}
In this part, we introduce the details of the compared baselines as follows:

\textbf{Textual methods.} We compare \method{} with two textual methods:

\begin{itemize}
    \item \textbf{SMILES2vec:}~\cite{goh2017smiles2vec} is a SMILES-based model that applies deep neural networks over tokenized SMILES sequences, using embedding layers followed by recurrent or convolutional layers to encode sequential chemical structure.
    \item \textbf{SMILES-BERT:}~\cite{wang2019smiles} is a SMILES-based model that leverages large-scale unsupervised pre-training using a BERT-style masked language modeling objective, followed by fine-tuning on downstream molecular property prediction tasks.
\end{itemize}

\begin{algorithm}[t]
\caption{{\method{}: Enhancing Molecular Property Prediction via Prototype-Guided Multimodal Learning}}
\label{alg:protomoL}
\begin{algorithmic}[1]
\Require Molecular graph $\mathcal{G}=(V,E,X)$, textual description $\mathcal{T}$, label $y$, 
hyperparameters $(N,K,\lambda_{align},\lambda_{pred},\lambda_{proto})$
\Ensure Trained parameters $\Theta$, prediction $\hat{y}$

\State \textbf{Initialization:} Initialize class-specific prototype set 
$\mathcal{P}=\{\mathbf{p}_1,\dots,\mathbf{p}_N\}$ with learnable parameters.

\For{each training step}
    \State \textbf{Graph Encoding:} 
    $\{\mathbf{z}_g^{(l)}\}_{l=1}^{L} \gets \mathrm{GNN}(\mathcal{G})$
    \State \textbf{Text Encoding:} 
    $\{\mathbf{z}_t^{(l)}\}_{l=1}^{L} \gets \mathrm{Transformer}(\mathcal{T})$
    \For{$l=1,\dots,L$}
        \State \textbf{Cross-Modal Interaction:}
        $(\tilde{\mathbf{z}}_g^{(l)},\tilde{\mathbf{z}}_t^{(l)}) 
        \gets \mathrm{CrossAttn}(\mathbf{z}_g^{(l)},\mathbf{z}_t^{(l)})$
        \State \textbf{Feature Fusion:} 
        $\hat{\mathbf{z}}_g^{(l)}=\mathbf{z}_g^{(l)}+\tilde{\mathbf{z}}_g^{(l)}$, 
        $\hat{\mathbf{z}}_t^{(l)}=\mathbf{z}_t^{(l)}+\tilde{\mathbf{z}}_t^{(l)}$
    \EndFor
    \State \textbf{Prototype Projection:}
    Map $\hat{\mathbf{z}}_g^{(L)}$ and $\hat{\mathbf{z}}_t^{(L)}$ into shared space 
    $\mathcal{P}$ to obtain similarity scores $\alpha_g$, $\alpha_t$.
    \State \textbf{Top-$K$ Selection:} 
    Retain the top-$K$ prototypes with highest activations for each sample.
    \State \textbf{Loss:}
    \[
        \mathcal{L}_{total}
        =\lambda_{pred}\,\mathcal{L}_{pred}
        +\lambda_{align}\mathcal{L}_{align}
        +\lambda_{proto}\,\mathcal{L}_{proto}.
    \]
    \State \textbf{Parameter Update:}
    $\Theta \leftarrow \Theta - \eta\nabla_{\Theta}\mathcal{L}_{total}$
\EndFor

\State \textbf{Prediction:} 
Given a new sample, from $\{\ \hat{\mathbf{z}}_g^{(i)}\}_{i=1}^{L}$ to obtain 
$\hat{\mathbf{o}}=\frac{1}{L}\sum_i \mathrm{Linear}^{(i)}(\hat{\mathbf{z}}_g^{(i)})$, 
and output 
\[
\hat{y}=
\begin{cases}
\arg\max_c \hat{\mathbf{o}}_c, & \text{(classification)}\\[4pt]
\hat{\mathbf{o}}, & \text{(regression)}.
\end{cases}
\]
\end{algorithmic}
\end{algorithm}

\textbf{Graph-based methods.} We compare \method{} with fourteen graph-based methods:

\begin{itemize}
    \item \textbf{GraphPT:}~\cite{CandA} proposes a general GNN pre-training framework to capture local and global structural information.
    \item \textbf{GraphSAGE:}~\cite{EdgePred} introduces an inductive framework for learning node embeddings by sampling and aggregating features from a node’s local neighborhood, enabling generalization to unseen nodes and efficient learning on large-scale graphs.
    \item \textbf{DGI:}~\cite{Infomax} proposes an unsupervised graph representation learning method by maximizing mutual information between local node embeddings and a global summary vector of the graph, using a contrastive objective to distinguish positive samples from corrupted negatives.
    \item \textbf{JOAOv2:}~\cite{JOAO} automates the selection of graph augmentations for contrastive learning by jointly optimizing augmentation probabilities and encoder parameters, enabling adaptive and effective pre-training without manual augmentation tuning.
    \item \textbf{GraphCL:}~\cite{GraphCL} introduces a framework that performs contrastive learning on graphs by applying random graph augmentations and maximizing agreement between augmented views of the same graph to learn expressive and generalizable graph-level representations.
    \item \textbf{GraphLoG:}~\cite{graphlog} proposes a self-supervised framework that captures both local and global structural patterns by designing a contrastive objective between instances sampled from the same graph and graphs sampled from the same class, encouraging representations to encode multi-scale topological semantics.
    \item \textbf{MICRO-Graph:}~\cite{micrograph} introduces a pre-training strategy that leverages biologically meaningful graph motifs as semantic priors. The method generates motif-aware positive samples and contrasts them against randomly corrupted negatives, enabling the learned representations to emphasize motif-relevant substructures and enhance downstream task performance.
    \item \textbf{MGSSL:}~\cite{mgssl} proposes a self-supervised framework that leverages chemical motifs as supervision. It detects frequent substructures in molecular graphs and applies contrastive learning to align motif-level and global representations, enhancing downstream property prediction.
    \item \textbf{GraphFP:}~\cite{graphft} introduces a self-supervised learning framework where molecules are decomposed into fragments. The model is pretrained to predict the presence and arrangement of these fragments, capturing meaningful substructure patterns. During finetuning, the learned fragment-aware representations are transferred to downstream molecular property prediction tasks.
    \item \textbf{GROVE:}~\cite{grove} proposes a transformer-based architecture pretrained using a combination of contextual property prediction and motif prediction tasks. It leverages graph-level and substructure-level objectives to capture both global and local chemical semantics, enabling effective transfer to molecular property prediction tasks.
    \item \textbf{SimSGT:}~\cite{simsgt} introduces a refined masked graph modeling approach that rethinks the design of tokenizers and decoders. It proposes a domain-aware graph tokenizer that generates discrete tokens from molecular graphs and a relation-aware decoder that reconstructs masked graph components, improving molecular representation quality.
    \item \textbf{MoAMa:}~\cite{moama} introduces a motif-guided masking strategy that incorporates chemical substructure (motif) information into the attribute masking process. By prioritizing the masking of functionally significant atoms and bonds, the method enhances pretraining effectiveness and guides the model to learn chemically meaningful representations.

    \item \textbf{Uni-Mol:}~\cite{Uni-Mol} proposes a unified framework for 3D molecular representation learning using SE(3)-equivariant positional encoding and a Transformer backbone. It is pre-trained on large-scale 3D molecular data via masked atom prediction and position denoising, supporting downstream tasks like property prediction and conformation generation.
    \item \textbf{S-CGIB:}~\cite{S-CGIB} introduces a pre-training framework where GNNs learn molecular representations through a subgraph-conditioned Graph Information Bottleneck (GIB). It generates subgraphs as conditionals and trains GNNs to preserve task-relevant information while compressing irrelevant features, improving generalization to downstream molecular tasks.
\end{itemize}

\textbf{Multimodal methods.} We compare \method{} with three multimodal methods that integrate textual and graph-based features:

\begin{itemize}
    \item \textbf{Tri-SGD:}~\cite{lu2024multimodal} proposes a fusion-based framework that integrates SMILES-based chemical language representations with molecular graph structures. It encodes SMILES using sequence models and graphs using GNNs, and then fuses the two modalities via concatenation followed by dense layers to jointly learn features for drug property prediction.
    \item \textbf{MMSG:}~\cite{wu2023molecular} introduces a dual-encoder framework where SMILES strings are processed using Transformer-based encoders and molecular graphs are encoded via GNNs. The model performs modality-specific encoding followed by joint embedding fusion through a shared representation space to enhance molecular property prediction.
    \item \textbf{MDFCL:}~\cite{gong2025mdfcl} proposes a contrastive learning framework that integrates SMILES and graph modalities. It uses dual encoders to extract representations from each modality and employs cross-modal contrastive objectives to align them, enhancing semantic consistency and improving prediction performance.
\end{itemize}

\subsection{Algorithm}

{We provide the overall process of \method{} in Algorithm~\ref{alg:protomoL}.
\method{}. The molecular graph $\mathcal{G}$ is encoded by a multi-layer GNN to capture hierarchical structural representations, while the textual description $\mathcal{T}$ is encoded through a Transformer to extract semantic information. At each layer, bidirectional cross-modal attention enables the interaction between graph and text embeddings, yielding refined representations that jointly preserve molecular topology and contextual semantics. The fused representations are projected into a shared prototype space $\mathcal{P}$, from which the top-$K$ activated prototypes are selected to guide prediction. The model is optimized end-to-end using a composite loss that combines the task-specific predictive loss $\mathcal{L}_{pred}$, the cross-modal alignment loss $\mathcal{L}_{align}$, and the prototype contrastive loss $\mathcal{L}_{proto}$. The final output $\hat{y}$ is obtained by aggregating the layer-wise graph representations and mapping them through the learned prototypes for classification or regression tasks.}

\subsection{Case-Level Visualization of Prototype Activations}\label{sec:vis_new}

{We performe a case-level visualization and analysis of prototype activations to interpret the learned molecular representations more clearly. We trained \method{} on the BACE dataset and randomly selected two inactive molecules ( in Figure.~\ref{fig:proto_vis} (a) and (c) ) and one active molecule ( in Figure.~\ref{fig:proto_vis} (b) ) for prototype-based prediction and visualization. In addition, we quantified the activation strength of each prototype within its corresponding molecule, where higher values indicate stronger prototype responses.}

As shown in Figure.~\ref{fig:proto_vis}, distinct prototypes are activated for different molecular classes. For the inactive molecules in Figure.~\ref{fig:proto_vis} (a) and Figure.~\ref{fig:proto_vis} (c), the activated prototypes correspond to simple aromatic or polar fragments such as benzene rings, amide linkages, and carbonyl groups, with Prototype P0 exhibiting the strongest activation. In contrast, the active molecule in Figure.~\ref{fig:proto_vis} (b) primarily activates prototypes associated with more complex and biologically relevant substructures, including cyclohexylamide, naphthalene systems, biphenyl scaffolds, and carbonyl bridges, with Prototype P6 showing the highest activation strength. These observations demonstrate that prototype activations differ substantially between active and inactive molecules and correspond to chemically meaningful structural motifs, highlighting both the interpretability and chemical significance of \method{}’s learned prototypes.

\begin{figure*}[t]
\centering
\includegraphics[width=1.0\textwidth]{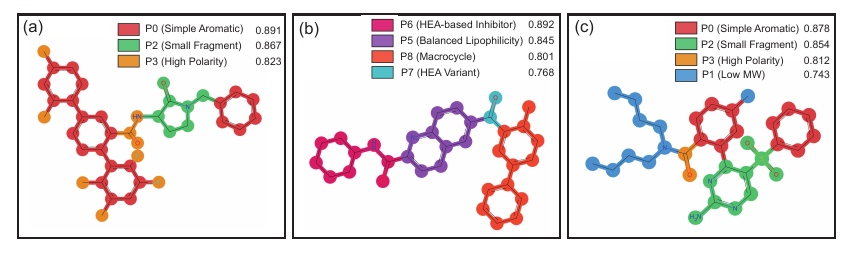}
\caption{{Visualization of prototype activations for active (b) and inactive (a,c) molecules on the BACE dataset.}}
\label{fig:proto_vis}
\end{figure*}
\end{document}